\documentclass[fleqn,11pt]{SelfArx} 

\usepackage{hyperref}
\hypersetup{
    hyperindex, linktoc=page, breaklinks,
    pdfauthor={Carlos Garcia-Saura},pdftitle={Central Pattern Generators for the control of robotic systems}
}

\usepackage{listings}

\definecolor{color1}{RGB}{0,0,90} 
\definecolor{color2}{RGB}{0,20,20} 

\JournalInfo{Independent Study Option report, April 2015}
\Archive{MSc in Computing (Spec. in Artificial Intelligence)\\ Imperial College London}
\PaperTitle{Central Pattern Generators \\for the control of robotic systems}
\PaperTitleNoLineBreaks{Central Pattern Generators for the control of robotic systems}
\Authors{Author: Carlos \textsc{Garcia-Saura}\\Supervisor: Prof. Murray \textsc{Shanahan}}
\newcommand{\keywordname}{Keywords}

\Abstract{
Bio-inspired control of motion is an active field of research with many applications in real world tasks. In the case of robotic systems that need to exhibit oscillatory behaviour (i.e. locomotion of snake-type or legged robots), Central Pattern Generators (CPGs) are among the most versatile solutions. These controllers are often based on loosely-coupled oscillators similar to those found in the neural circuits of many animal species, and can be more robust to uncertainty (i.e. external perturbations) than traditional control approaches.

This project provides an overview of the state-of-the-art in the field of CPGs, and in particular their applications within robotic systems. The project also tackles the implementation of a CPG-based controller in a small 3D-printed hexapod.
}

\begin{document}

\flushbottom 

\maketitle 

\tableofcontents 

\thispagestyle{empty} 


\section*{Introduction}

\addcontentsline{toc}{section}{Introduction}

We live in an exciting time for robotics.
The computing revolution that made powerful processing technology widespread, has contributed to the rise of domestic 3D printers, and being able to manufacture customized structures at a low cost, actuate them with off-the-shelf hardware, and control them with computers, has proven to be a winning combination for robotics. New designs keep appearing each day, in many occasions taking inspiration from nature; this is the case of snake-type or legged robots, among others.

But biologically-inspired robots often have many degrees of freedom that present as a true challenge for traditional control approaches (i.e. centralized systems that compute motion patterns globally and control an articulated robot at the joint level)\cite{ijspeertCPGreview}.
Whilst it is possible to solve some of those problems for controlled environments, many algorithms are simply not prepared for the real world\cite{ekeberg93,HooperCPGs00}.
It is here where neuroscience is answering many questions, and we are starting to see real world robots that not only look like, but also interact with the environment just as living creatures do.

The mechanisms behind higher level motion planning in animals -cognition- have been a subject of research for many decades, but the exact way in which the brain operates is yet a subject of speculation.
What we know is that evolution has developed efficient ways to abstract very complex locomotion patterns, and turn them into simple control signals that can be effortlessly handled by the brain\cite{ijspeertCPGreview,ekeberg93,HooperCPGs00}.
Given the intrinsic periodicity of locomotion, it is reasonable to think that neural circuits do not need to explicitly control the position of every articulation, but rather modulate higher level parameters such as the amplitude or frequency of each gait\cite{ijspeertSalamander,HerreroBioInspiredStrategiesModular11}.

Often involved in motion control, Central Pattern Generators (CPGs) are biological neural networks that produce rhythmic output patterns without the need for sensory feedback.
This means that either individual neurons or the way they are connected to each other, can lead to oscillatory dynamics that are useful for various biological functions (i.e. walking gaits, respiration, or even circadian rhythms).
These oscillatory neural circuits have been studied in many animals such as
stick insects and cockroaches\cite{simNeuralDynRoboticLeg}, sea angels\cite{clione98}, fish\cite{ekeberg93,crespiFish08} and other vertebrates such as the salamander\cite{ijspeertSalamander}, cats\cite{lynxRobot14} and mice\cite{neuromechTetrapodCoord}.
In the case of vertebrates, CPG circuits involved in locomotion are often situated in the spinal cord.

These pattern-generating neural networks as observed in biology have inspired roboticists, and it has already been possible to replicate locomotion behaviours in many robot designs. Some examples are the salamander-inspired \emph{Pleurobot}\cite{ijspeertSalamander} from EPFL, quadrupeds like \emph{Lynx}\cite{lynxRobot14} from the same university, the worm-like \emph{Cube Revolutions}\cite{HerreroBioInspiredStrategiesModular11} from Univ. Autonoma de Madrid, or the stick-insect robot \emph{Hector}\cite{hector12} from Bielefeld University.


\section{Computational Models of CPGs}

Controllers based on central pattern generators can indeed be very robust against uncertainties in the environment (i.e. autonomous stepping over an obstacle that appears in the way)\cite{herreroNonLinearMatsuoka10},
but whenever a robot needs to interact with the real world, delays in its response
may still have dramatic consequences.
Thus, real time decision-making is crucial for these sort of interactions, so detailed simulations of biologically realistic CPGs (i.e. spiking networks of Hodgkin-Huxley neurons) are frequently deemed too computationally intensive for robot control.
Rather, simplified models of CPGs have been developed in order to facilitate their real-world implementation and to meet those strict timing requirements.

\subsection{Biological Neural Networks}
In this section we are going to focus on CPG implementations that remain close to biology by modelling the dynamics of neural networks for motion control.

\begin{figure}[ht]\centering
\includegraphics[width=0.85\linewidth]{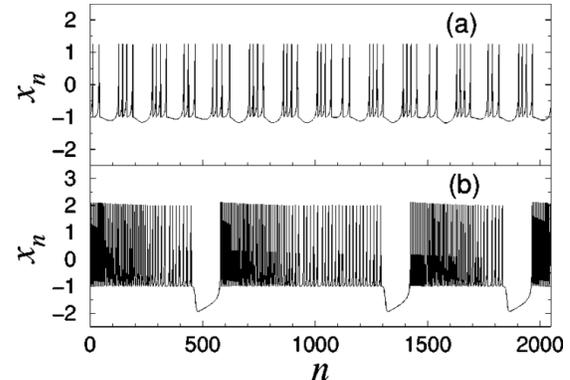}
\caption{Typical wave forms of the spiking-bursting behaviour generated by the \emph{Rulkov map} for two different sets of parameters. From \cite{rulkov02} by N.F. Rulkov.}
\label{fig:rulkov}
\end{figure}

The first model that we are going to study is the \emph{Rulkov map} type neuron. Whilst they are more efficient to compute than continuous time models, these neurons can still exhibit rich spiking dynamics (see Fig. \ref{fig:rulkov}) that allow for implementations of central pattern generators that closely resemble data from experiments with real ganglia\cite{rulkov02}.

\begin{figure}[ht]\centering
\includegraphics[width=0.8\linewidth]{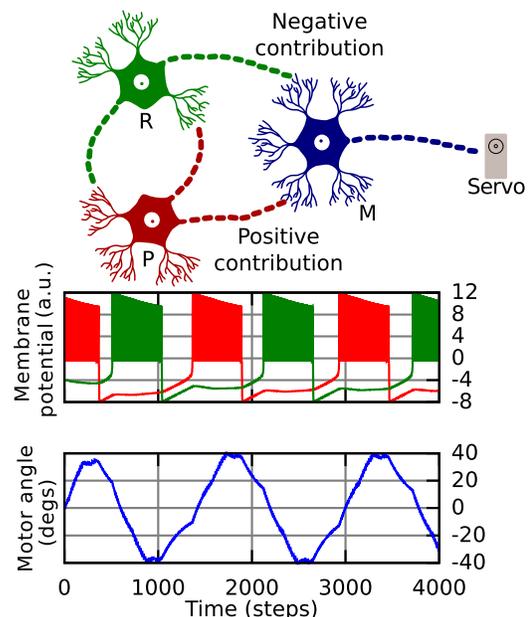}
\caption{CPG half-centre implemented with the Rulkov map model.
Neurons $R$ and $P$ negotiate a common rhythm through mutual inhibition, which results in anti-phase synchronization. The motoneuron $M$ produces a control signal (blue), an oscillation of the controlled joint.
Reprint from \cite{carron2011novel} by F. Herrero Carr\'{o}n.}
\label{fig:rulkovhalfcenters}
\end{figure}

The basic unit of a biological CPG is the \emph{half-centre}, which consists in two neurons (or populations) that cannot generate rhythmic activity by themselves, but may oscillate when coupled reciprocally (see Fig. \ref{fig:rulkovhalfcenters}).

A half-centre is said to work in \emph{escape} mode if the off neuron turns on after \emph{escaping} from inhibition, and
similarly, the \emph{release} mode means that the neurons turn on after \emph{being released} from inhibition.
Variations in the properties of each synapse can produce very different coalition results in a broad range: from the complete synchrony of firings (in counter-phase or coherent phase) to completely decoupled oscillations\cite{ijspeertCPGreview,ekeberg93,
HooperCPGs00,carron2011novel,
herreroDynamicalInvariants10}.

\begin{figure}[ht]\centering
\includegraphics[width=0.68\linewidth]{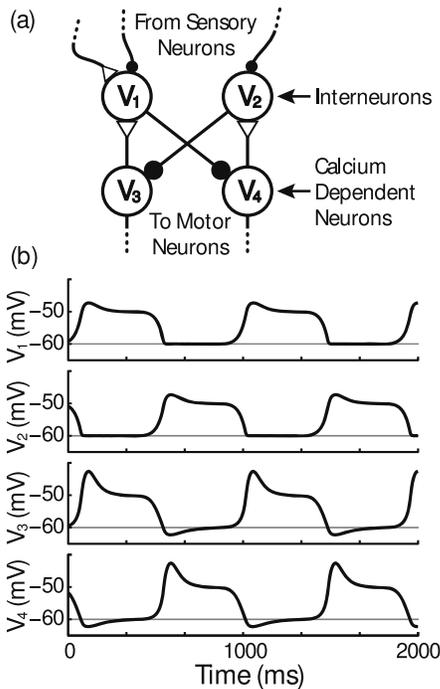}
\caption{(a) Two coupled half-centres, for the control of flexion ($V_3$) and extension ($V_4$) of one of the joints in a robot leg.
Excitatory synapses are represented as white triangles, and inhibitory synapses with black circles.
Output wave forms (b) were obtained with the \emph{leaky integrator} neuron model. The nature of the sensory input at $V_1$ and $V_2$ will be studied in further sections.
Reprint from \cite{simNeuralDynRoboticLeg} by M.A. Klein et al.}
\label{fig:halfcenters}
\end{figure}

In legged systems, for example, flexion and extension of a joint should never occur simultaneously. We can achieve this by using pairs of CPG half-centres coupled in anti-phase to control them (see Fig. \ref{fig:halfcenters}).
As in nature, flexor and extensor muscles alternate to achieve full oscillatory patterns.

We have already seen the \emph{Rulkov map}, but another neuron model that has been used in plenty of research in this field is the \emph{leaky integrator} type. This non-spiking model is based on a set of differential equations that are
capable of representing either individual non-spiking inter-neurons or the firing rate of a population\cite{simNeuralDynRoboticLeg,neuromechTetrapodCoord,MantisModelPostures}.

\subsubsection{Controlling locomotion with synaptic gating}
\begin{figure}[ht]\centering
\includegraphics[width=0.65\linewidth]{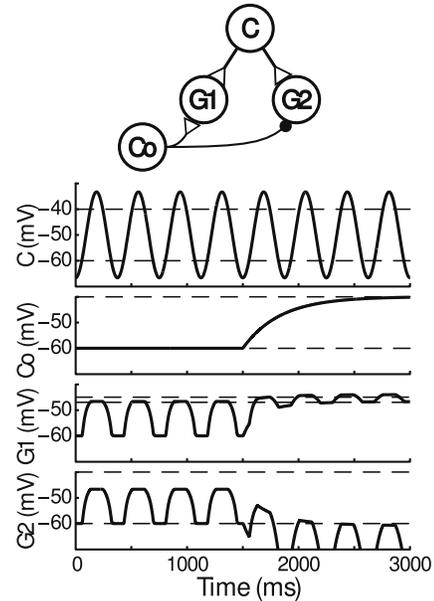}
\caption{Gating effect of a context neuron $C_o$.
$C$ is a control neuron that oscillates.
Synaptic conductances can be set in such way that $C_o$ controls whether neurons $G_1$ and $G_2$ operate within or outside their range, hence switching their ability to propagate the activity from $C$ any further.
In this case only $G_2$ can conduct when $C_o$ is silent. But when $C_o$ becomes active, it excites $G_1$ (allowing it to conduct) and inhibits $G_2$.
Reprint from \cite{simNeuralDynRoboticLeg} by M.A. Klein et al.}
\label{fig:gating}
\end{figure}

Coupling between CPGs is often achieved with the direct excitation/inhibition of neurons between each oscillatory source.
Synaptic gating, however, allows for richer interactions, as it provides neurons with the ability to dynamically alter the strength of some other synapse within the network.
This means that it can either facilitate or suppress the communication between two other neurons. An example implementation that demonstrates this effect is shown in Figure \ref{fig:gating}.

These methods make it possible to achieve a wide range of modulations in the locomotive patterns produced by a biological neural network\cite{ijspeertCPGreview,ekeberg93,simNeuralDynRoboticLeg}:

\begin{itemize}[noitemsep]
\item The frequency of oscillation can be controlled by varying the input current for neurons within a CPG half-centre.
\item Gating the output of the CPG allows for amplitude-controlled oscillations.
\item Also, an asymmetric modulation of the couplings between two half-centres can produce variations of the duty cycle.
\end{itemize}

There are many examples of real-world robots that use CPG implementations based on biological neural networks.
The \emph{Rulkov map} neuron model has been demonstrated for the control of snake-type robots\cite{carron2011novel,rulkovUnderwater10} as well as in robots with limited wheels that move by means of oscillations\cite{UrziceanuCPGdifferential11}.
The \emph{leaky integrator} neuron model has been used in robot controllers for quadrupeds\cite{neuromechTetrapodCoord} and insects\cite{MantisModelPostures}.

Many other biological models are available\cite{ijspeertCPGreview,HooperCPGs00}, and incorporating richer dynamics such as \emph{decaying synapses} to these networks can also be useful to better replicate what is observed in nature\cite{simNeuralDynRoboticLeg}.

\subsection{Coupled oscillator networks}
Another approach towards the implementation of central pattern generators in robots is to move away from biology, and make use of mathematical models of synchrony such as the \emph{Kuramoto} oscillators.
Rather than trying to explain rhythmogenesis, these models focus on the study of the effects of inter-oscillator couplings, and assume the presence of underlying oscillatory mechanisms.

\begin{figure}[ht]\centering
\includegraphics[width=0.89\linewidth]{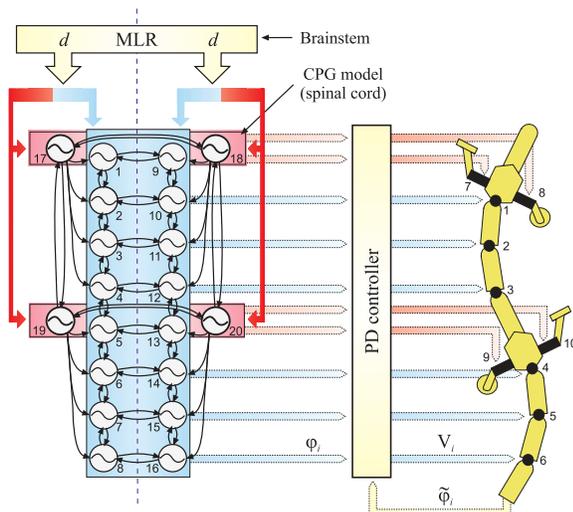}
\caption{Connectivity within the controller of \emph{Salamandra robotica}.
The arrangement of oscillators mimics the spinal cord of a lamprey\cite{ekeberg93} where salamander limbs have been added.
Sensory feedback is not incorporated into the CPG in this case. Instead, each actuator has a PD controller that ensures an accurate motion.
Reprint from \cite{ijspeertSalamander} by A.J. Ijspeert et al.}
\label{fig:cpgsalamander}
\end{figure}

\begin{figure}[ht]\centering
\includegraphics[width=\linewidth]{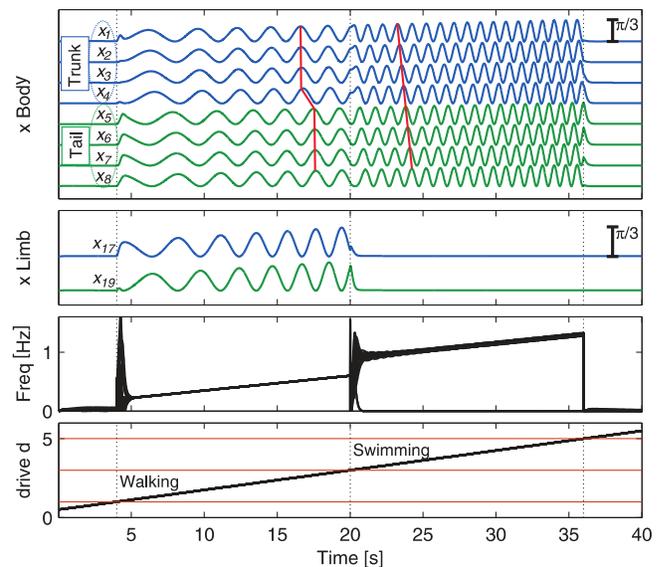}
\caption{Gait transition from walking to swimming in \emph{Salamandra robotica}, for one side of the spinal cord.
A walking gait arises at low levels of drive.
When limb oscillators saturate, the swimming gait takes over control.
The red vertical markers show both a standing wave for walking and a traveling wave for swimming.
Reprint from \cite{ijspeertSalamander} by A.J. Ijspeert et al.}
\label{fig:salamanderTransition}
\end{figure}

Even though the models are simple, it is still possible to replicate many of the characteristics that appear in nature.
For instance, \emph{Salamandra robotica} from EPFL (see Figs. \ref{fig:cpgsalamander} \& \ref{fig:salamanderTransition}) successfully demonstrated a smooth transition between the walking and swimming gaits\cite{ijspeertSalamander}.

The controller of \emph{Salamandra robotica} is very interesting indeed, as it was built upon four biological hypotheses:
\begin{enumerate}[noitemsep]
\item The salamander has a body CPG that spontaneously produces traveling waves (swimming gait) such as the one produced by the lamprey\cite{ekeberg93} when activated with a tonic drive signal.
\item Limb CPGs have lower intrinsic frequencies than the segments or body oscillators.
\item When the limb CPG is activated, it forces the whole system into walking mode. To achieve this, $limb \rightarrow segment$ couplings are kept stronger than $segment \rightarrow segment$ interconnections.
\item Limb CPGs cannot oscillate at high frequencies, so they saturate at high levels of drive. The natural swimming gait can then take over control.
\end{enumerate}

Altogether, the resulting controller can efficiently actuate two different gaits in a 10 DOF robot with only one input signal (see Fig. \ref{fig:salamanderTransition}).
The direction of locomotion can also be controlled with an additional input signal that modulates the oscillations of each side of the spinal cord.

\begin{figure}[ht]\centering
\includegraphics[width=1.02\linewidth]{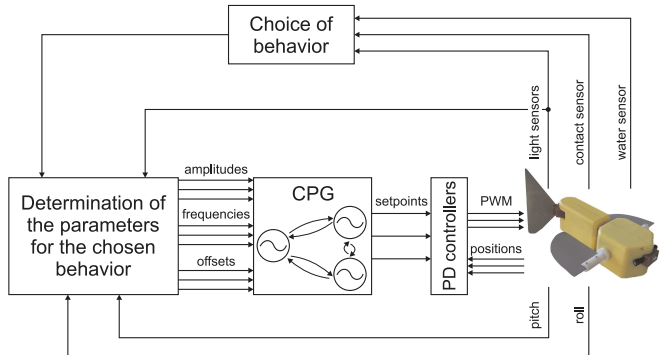}
\caption{Diagram of the CPG-based controller for a 3 DOF fish robot.
The three oscillatory centres control one caudal and two pectoral fins. Sensory feedback will be further discussed in the next section.
Reprint from \cite{crespiFish08} by A. Crespi et al.}
\label{fig:fishcpg}
\end{figure}

For this study we wanted to better understand how these oscillator networks worked at a lower level, but the ten degrees of freedom of the salamander robot discouraged a re-implementation.
Instead, we found another project from the same laboratory: a fish robot with three degrees of freedom (see Fig. \ref{fig:fishcpg}).

\begin{figure}[ht]\centering
\includegraphics[width=0.9\linewidth]{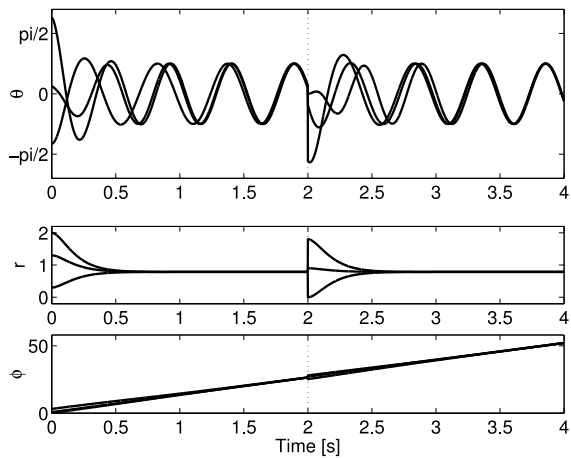}
\caption{CPG limit cycle for the fish robot.
The upper panel shows the output wave forms for the three reciprocally-coupled oscillators. Following random initial conditions, the system quickly stabilizes into coherent synchrony.
The steady state is also rapidly reached after adding random perturbations to the state variables ($r$,$\phi$,$x$) at $t=2s$.
Reprint from \cite{crespiFish08} by A. Crespi et al.}
\label{fig:fishcpglimitcycle}
\end{figure}

The CPG model for the fish robot is indeed much more simple than the one of the salamander, but on the other hand it can show in a more intuitive way basic concepts such as oscillation coupling and rhythm negotiations (see Fig. \ref{fig:fishcpglimitcycle}).

Among other robots that use mathematical models of synchrony in the form of coupled oscillator networks are quadrupeds like \emph{Lynx}\cite{lynxRobot14} also from EPFL, and
\emph{Pneupard}\cite{CPGstretchReflexPneumaticQuadruped} from Osaka University.

\subsection{Incorporation of sensory feedback}

In the case of the fish robot, sensory feedback was incorporated by directly modulating the parameters within the CPG according to predefined gaits.
For instance, the fish robot has two light intensity sensors, one on each side, that were used to achieve a photo-taxis behaviour with a Braitenberg vehicle approach\cite{crespiFish08} (the oscillation of each pectoral fin is modulated by the light intensity perceived in the opposite side; this has the effect of the robot turning towards light sources in the environment).

Whilst such an approach can work well for simple neural networks, a more interesting and biologically realistic method would incorporate sensory information by encoding system-wide behavioural reflexes into the network itself.

For instance, lets recall that both the salamander and fish robot rely on PD controllers that ensure that each motor joint reaches a target angle, calculated by the pattern-generating network.
Those PD controllers are indeed using information from rotation encoders at the joint level, so why not directly incorporate that information into the CPG instead?
In nature, this simple form of sensory feedback is at the muscular level, where \emph{strech} or \emph{edge} cells can sense muscle contractions,
and feed them back into the network.
In humans, these cells play a very important role in the spatial perception of the body (proprioception).
This approach is also much more robust against uncertainty in external perturbations than open loop CPG controllers.

\begin{figure}[ht]\centering
\includegraphics[width=0.9\linewidth]{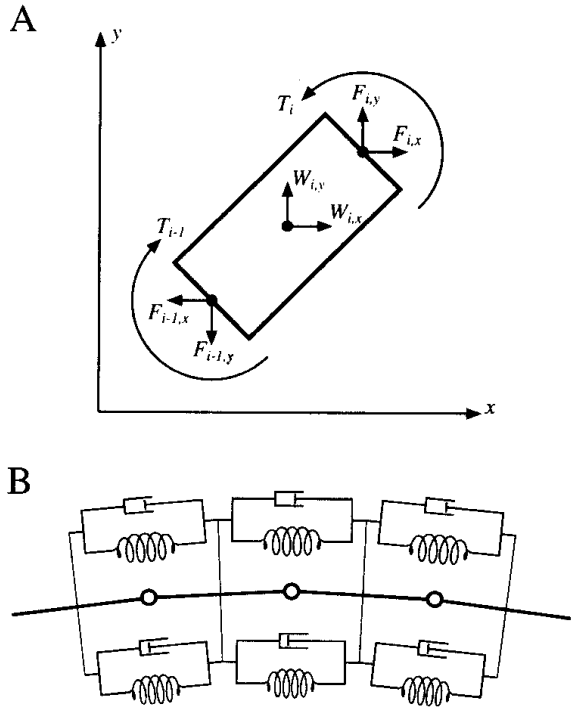}
\caption{For a lamprey-like robot, \textbf{A} shows the forces acting on link $i$: muscular torques $T_i$ and $T_{i-1}$, water forces $W_i$, and inner forces from neighbouring links $F_i$ and $F_{i-1}$. In \textbf{B} we can see the schematic representation of three joints. Each spring-dampener assembly is used to model a single muscle and its interaction with water forces.
The spring constants are changed by the activity of the corresponding motoneuron in the CPG.
With this model, it is possible for the network to change both the total bending force and the local stiffness of each segment in the body.
Reprint fom \cite{ekeberg93} by O. Ekeberg.}
\label{fig:neuromechanical}
\end{figure}

However, until very recently this feedback concept could not be exploited in real robots due to the elevated cost of high precision rotatory encoders.
Instead, virtual stretch cell models (such as the \emph{linear Hill muscle model}) were developed in order to achieve realistic \emph{neuro-mechanical} simulations. This concept is discussed in Figure \ref{fig:neuromechanical}.

Other simple forms of sensory feedback are \emph{foot contact} sensors. Recent studies have shown that hair fibres present in the feet of rats (\emph{carpal vibrissae}) behave as contact sensors that influence the kinematics of locomotion\cite{CarpalVibrissaeRats}. Whilst the incorporation of such feedback into CPG networks is far from trivial, these kind of sensors have the advantage of greater simplicity and lower implementation cost than rotatory encoders.

\pagebreak

The last sensory input related to locomotion that deserves a mention is the vestibular system. Placed in the inner ear of most mammals, it plays an essential role in balance. One of the best examples is bipedal walking, where equilibrium perception is seamlessly incorporated into the neural mechanisms responsible for motion control.
CPG-based robot controllers can also benefit from this form of input by using off-the-shelf accelerometers.

As a final remark for this section, we must say that CPGs often have many parameters that need to be finely adjusted in order to achieve an efficient locomotion. The automatic entrainment of these models is still a subject of research\cite{onlineLearningCPGmodular,
herreroFlexEntrainment11,
LiuCPGworkspaceTrajectoryAdaptiveLocom11}.

Though there is plenty of software available for the simulation of robots, only few combine environment physics with neuron dynamics.
The free version of \emph{AnimatLab}\footnote{\url{http://animatlab.com/}} provides a nice toolbox for basic experimentation, but more powerful open-source tools such as \emph{neuroConstruct}\footnote{\url{http://www.neuroconstruct.org/}} and \emph{geppetto}\footnote{\url{http://www.geppetto.org/}} are being developed thanks to the efforts of the \emph{OpenWorm}\footnote{\url{http://www.openworm.org/}} project.

\clearpage

\newpage

\section{Experiments with a 3 DOF hexapod}

\begin{figure}[ht]\centering
\includegraphics[width=\linewidth]{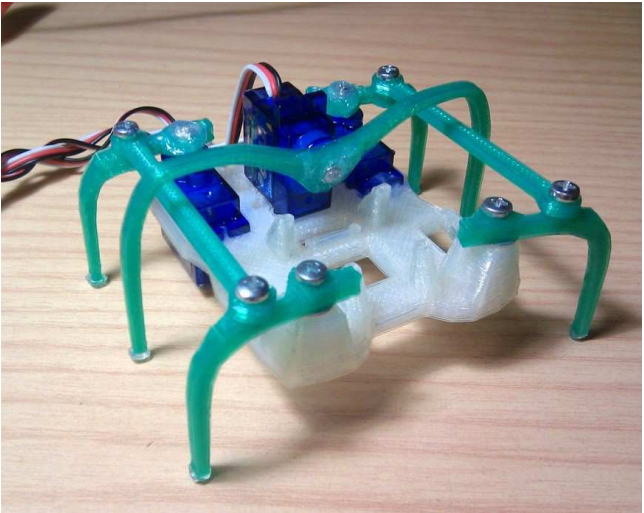}
\caption{Micro-hexapod used for the experiments. 
The robot is composed of eight 3D-printed parts (in white and green), nine screws, and three servomotors (blue).
The three degrees of freedom consist in two pairs of mechanically-coupled legs at each side, and also a middle joint that selects which of the sides makes ground contact.}
\label{fig:hexapod}
\end{figure}

For the experimental part of this study, we decided to implement a CPG controller for a small 3D-printed hexapod robot\footnote{Originally designed by Ijon Tichy: \\ \url{http://www.thingiverse.com/thing:5156/}} (see Fig. \ref{fig:hexapod}).
Since it has three degrees of freedom, it was the perfect candidate for a re-implementation of the controller presented in \emph{``Controlling swimming and crawling in a fish robot using a central pattern generator''} by A. Crespi et al.\cite{crespiFish08}

Our motivation for using the fish model as a reference implementation was that its oscillatory network model is very similar to the one employed in \emph{Salamandra robotica}\cite{ijspeertSalamander} from EPFL.
With three degrees of freedom instead of ten, the fish robot seemed like a better choice as a didactic exercise.

The CPG is built upon amplitude-controlled phase oscillators derived from the Kuramoto model of synchrony. Next, we will show the set of differential equations that define the network:

\begin{equation}
\dot{\phi_i} = \omega_i + \sum\limits_{j}{[\omega_{ij} r_j \sin{(\phi_j-\phi_i-\varphi_{ij})}]}
\label{eq:phidoti}
\end{equation}
\begin{equation}
\ddot{r_i} = a_r (\frac{a_r}{4} (R_i-r_i)-\dot{r_i})
\label{eq:rddoti}
\end{equation}
\vspace{-5mm}
\begin{equation}
\ddot{x_i} = a_x (\frac{a_x}{4} (X_i-x_i)-\dot{x_i})
\label{eq:xddoti}
\end{equation}
\vspace{-5mm}
\begin{equation}
\theta_i = x_i+r_i \sin{(\phi_i)}
\label{eq:thetai}
\end{equation}

Where:
\begin{itemize}[noitemsep]
\item $\theta_i$ is the output angle of each oscillatory centre
\item $\phi_i$, $r_i$ and $x_i$ are the state variables that encode the variation in time of every phase, amplitude and offset
\item The control parameters for each oscillator are $\omega_i$ (natural frequency), $R_i$ (target amplitude) and $X_i$ (target offset).
\item $a_r$ and $a_x$ are constant positive gains for Eqns. (\ref{eq:rddoti}) and (\ref{eq:xddoti}). In our case: $a_r = a_x = 2\ rad/s$.
\item Finally, $\omega_{ij}$ and $\varphi_{ij}$ are respectively the coupling weights and phase biases that determine how oscillator $j$ influences oscillator $i$.
\end{itemize}

The only change we made to the original equations in \cite{crespiFish08} was to replace the cosine function in Eqn. (\ref{eq:thetai}) with a sine. That way, the reference positions of the joints in the hexapod robot -corresponding to zero offset- match with the lateral segments being perpendicular to the body, and with a flat position of the middle segment (the same posture as in Fig. \ref{fig:hexapod}).

Equations (\ref{eq:rddoti}) and (\ref{eq:xddoti}) are critically-dampened second-order differential equations with stable fix points $R_i$ and $X_i$ respectively. With them it is possible to smoothly modulate the amplitude and offset of the oscillations in real time.

\begin{figure}[ht]\centering
\includegraphics[width=0.8\linewidth]{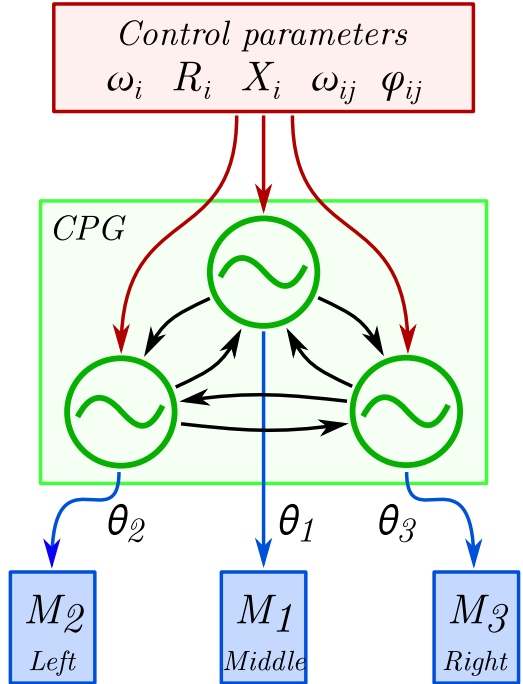}
\caption{Diagram of the CPG-based controller implemented for the hexapod robot.
A high level layer (red) modifies the control parameters in real time in order to determine the gait.
The three oscillatory centres (green) are coupled with constant weights.
Their output $\theta_i$ is fed into the PID controller of each servomotor.}
\label{fig:cpghexapod}
\end{figure}

The three oscillators $i=1,2,3$ correspond to the middle, left and right joints of the robot respectively, for which the maximum oscillation amplitudes are $A_{middle}=12\,^{\circ}$, $A_{left}=A_{right}=40\,^{\circ}$.
A schematic diagram with the employed topology is shown in Figure \ref{fig:cpghexapod}.

As in Crepi's article, each of the three oscillators were set to the same natural frequencies ($\omega_i=\omega$). Initially, the coupling parameters were set to:

\vspace{2mm}

$\omega_{ij}=0.5\ [1/s]\ \forall i \neq j$,
$\omega_{ii}=0.0\ [1/s]$,

$\varphi_{ij}=0.0\ [rad/s]\ \forall i \neq j$,
$\varphi_{ii}=0.0\ [rad/s]$.

\vspace{2mm}

In other words: the network is fully connected, but there are no self-couplings, and (originally) no phase biases.
With these parameters, the system converges into a regime where the phases (state variables $\phi_i$) grow linearly with a common rate $\omega$ and zero phase difference between the oscillators.

For the robot hexapod, gait information would need to be encoded into the phase biases $\varphi_{ij}$. Since the original paper was instead modulating the target offsets $X_i$ and amplitudes $R_i$, our model used $R_i$ exclusively as a smooth transition between the static state (zero amplitude or $R_i=0\,^{\circ}$) and motion, and $X_i$ was not necessary.

With this topology it is possible to define the following walking gaits:

\begin{itemize}[noitemsep]
\item \textbf{Forward walking} is achieved by setting

$\omega_{21}=\omega_{31}=0.5\ [1/s]$,
$\omega_{ij}=0.0\ [1/s]$ otherwise,

$\varphi_{21}=\pi/2\ [rad]$, 
$\varphi_{31}=-\pi/2\ [rad]$, 

\item \textbf{Backward walking} simply requires a shift in the phase couplings:

$\omega_{21}=\omega_{31}=0.5\ [1/s]$,
$\omega_{ij}=0.0\ [1/s]$ otherwise,

$\varphi_{21}=-\pi/2\ [rad]$, 
$\varphi_{31}=\pi/2\ [rad]$, 

\item \textbf{Clockwise rotation} is achieved with

$\omega_{21}=\omega_{31}=0.5\ [1/s]$,
$\omega_{ij}=0.0\ [1/s]$ otherwise,

$\varphi_{21}=\pi/2\ [rad]$, 
$\varphi_{31}=\pi/2\ [rad]$, 

\item Finally, \textbf{counter-clockwise rotation} uses

$\omega_{21}=\omega_{31}=0.5\ [1/s]$,
$\omega_{ij}=0.0\ [1/s]$ otherwise,

$\varphi_{21}=\pi/2\ [rad]$, 
$\varphi_{31}=\pi/2\ [rad]$, 
\end{itemize}

\subsection{Implementation of the CPG with Arduino}
The robot is controlled using a a ZUM BT-328 board\footnote{\url{http://www.bq.com/gb/placa-zum-bt/}}, which is based on the Arduino\footnote{\url{http://www.arduino.cc/}} platform.
The ordinary differential equations of the oscillator model were implemented using the \emph{Euler} method\footnote{\url{https://en.wikipedia.org/wiki/Euler_method}}, with the following update rule:

\begin{lstlisting}[language=C]
  for(int i=0; i<N; i++) {
    float sA_d = cW[i];
    for(int j=0; j<N; j++)
       sA_d += w[i][j] * sr[j] *
         sin(sA[j] - sA[i] - a[i][j]);
    new_sA[i] = sA[i] + T * sA_d;
    
    float sr_dd = ar * ( (ar/4) *
      (cR[i]-sr[i]) - sr_d[i] );
    new_sr_d[i] = sr_d[i] + T * sr_dd;
    new_sr[i] = sr[i] + T * sr_d[i];
    
    float sx_dd = ax * ( (ax/4) *
      (cX[i]-sx[i]) - sx_d[i] );
    new_sx_d[i] = sx_d[i] + T * sx_dd;
    new_sx[i] = sx[i] + T * sx_d[i];
  }
  
  for(int i=0; i<N; i++) {
    sA[i] = new_sA[i];
    sr[i] = new_sr[i];
    sx[i] = new_sx[i];
    sr_d[i] = new_sr_d[i];
    sx_d[i] = new_sx_d[i];
    output[i] = sx[i] + sr[i] *
      sin(sA[i]);
  }
\end{lstlisting}

Where $T$ is the time step for the Euler method. In our setup, we used a value of $T=0.01s$.


\section{Results and Discussion}

The hexapod implementation was a very useful test-bed for its simplicity, and we could experiment with a wide variety of parameters. For instance, we set $\varphi_{23}=\varphi_{32}=\pi\ rad$ so that the legs at both sides of the robot negotiated a counter-phase rhythm. Locomotion was then achieved by coupling one of the sides uni-directionally with the middle segment ($\omega_{12}=0.5\ [1/s]$, $\omega_{12}=0.0\ [1/s]$, $\varphi_{12}=\pi/2\ [rad]$).

\begin{figure*}[ht]\centering
\includegraphics[width=\linewidth]{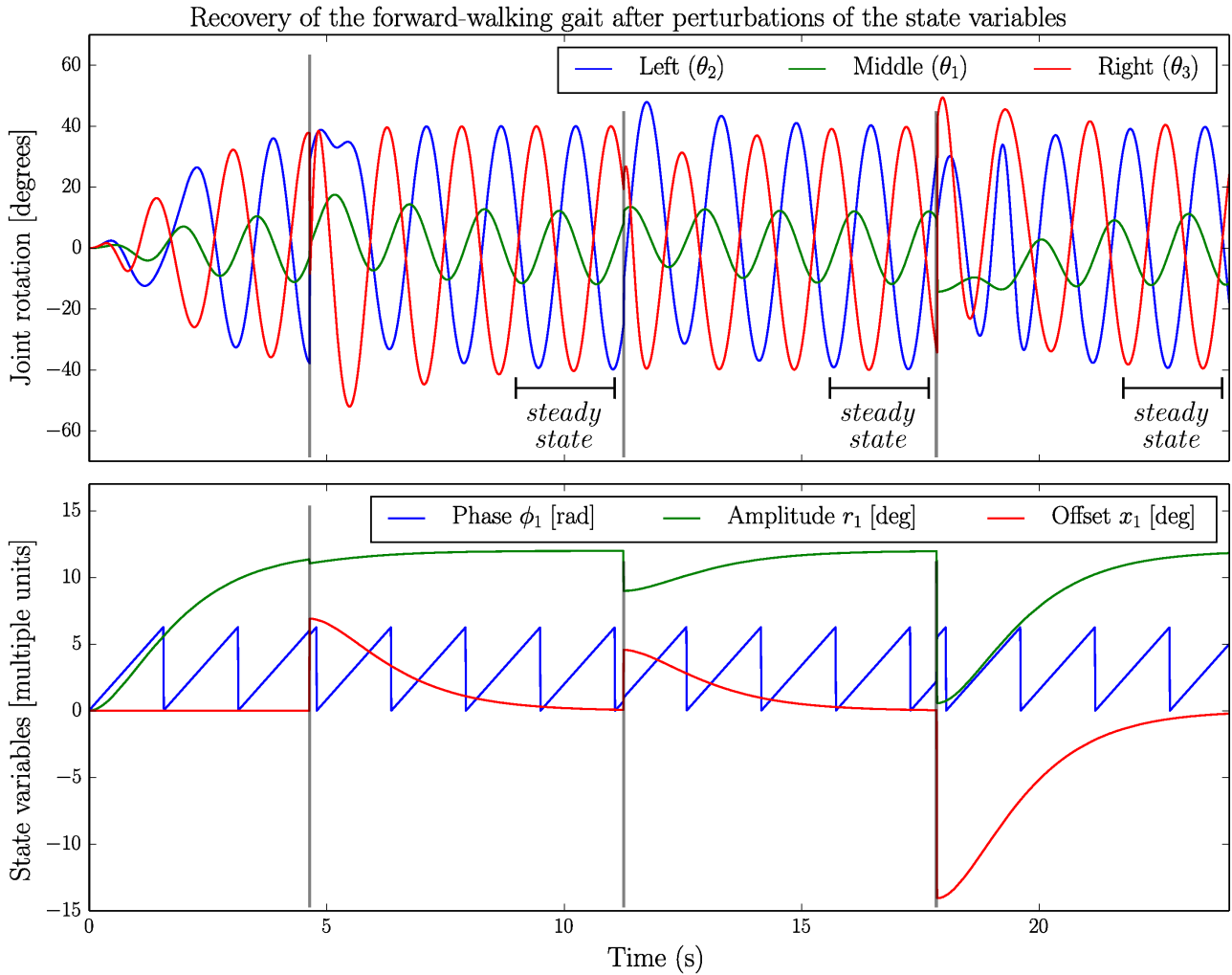}
\caption{The graph shows how the CPG network is capable of recovering from random perturbations. The upper panel displays the CPG output (angles for each motor joint). The state variables for the central joint are plotted in the lower panel. Vertical markers have been placed to signal the points where the perturbations were introduced; these would correspond to interactions with the environment (i.e. an obstacle in the way).}
\label{fig:recovery}
\end{figure*}

Another effect that we evaluated was the ability of the CPG to synchronize even when each oscillator had a different natural frequency. In a centralized approach with no couplings, slight variations of the intrinsic frequencies can lead to inconsistency in the gaits over time, due to phase drift.
We could observe instead how the CPG corrected those slight phase differences seamlessly in real time, even after perturbations had been applied to the state variables (see Fig. \ref{fig:recovery}).

\subsection{Smooth transitions between gaits}

We also tried to achieve a similar gait transitioning effect as the one in the salamander robot\cite{ijspeertSalamander}. By setting different natural frequencies for each oscillator, our intention was to create multiple resonant states (gaits) that arised at different levels of drive. At low frequencies the robot would for instance rotate in place, but upon a frequency increase a new forward gait would be negotiated among oscillators. Unfortunately we were not able to make this work, so more experimentation (i.e. the incorporation of additional oscillators into the network) would be needed.

\begin{figure*}[ht]\centering
\includegraphics[width=\linewidth]{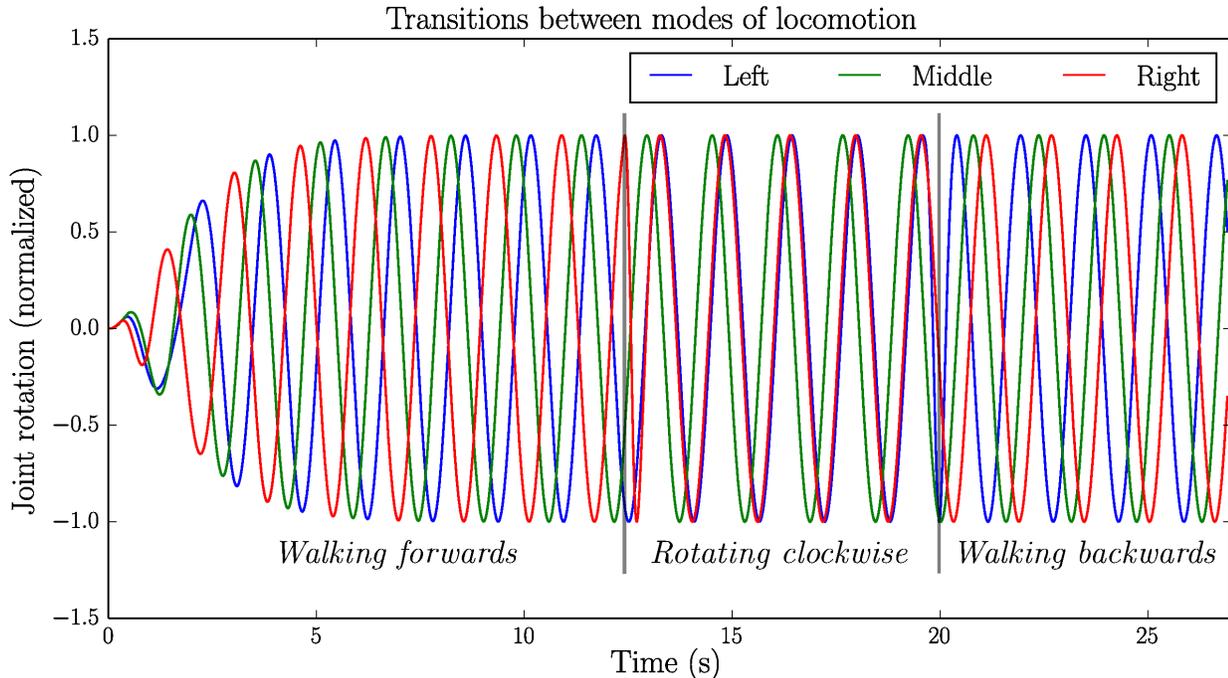}
\caption{Transitions between modes of locomotion for the 3 DOF hexapod robot.
The CPG-based controller was used in order to achieve smooth convergence into the desired pattern after each change of parameters.
The vertical bars at $t=12s$ and $t=20s$ show the instants when locomotion parameters were updated.
The amplitudes of each CPG have been normalized with the following values: $A_{left}=A_{right}=40\,^{\circ}$, $A_{middle}=12\,^{\circ}$.}
\label{fig:transitions}
\end{figure*}

Such a topologically-based control approach would not only be more realistic in biological terms, but also more interesting in regards to sensory input management.
For instance, joint-level feedback would allow to detect collisions with obstacles in the ground, eliciting obstacle-avoidance behaviour and thus producing a more efficient locomotion.

Rather than using a topological control approach, we achieved smooth transitions between each gait using the same method as in the fish robot paper\cite{crespiFish08}.
It basically employs the stability properties of CPG networks to automatically negotiate oscillations after abruptly changing their coupling parameters (see Fig. \ref{fig:transitions}).
This is a very simple method, but it demonstrates the flexibility of CPGs for both high-level and low-level motor controller implementations.

\subsection{Concluding remarks}

Evolution has developed biological neural networks with pattern-generating properties (CPGs) as an abstraction that facilitates efficient locomotive behaviour in animals. Scientific research has later developed methods to employ such abstractions in real-world problems.

This project has studied the applications of artificial central pattern generators in the field of robotics.
The state-of-the-art has been reviewed, and the most useful techniques have been shown, together with examples of their applications in real world robots.
The project has also tackled the implementation of a CPG-based controller in an hexapod with 3 degrees-of-freedom, and successfully demonstrated its ability to smoothly transition between various walking gaits. The robustness of the CPG has also been evaluated against random perturbations in the environment (c.f. Figs. \ref{fig:recovery} \& \ref{fig:transitions}).

Bio-inspired controllers based on central pattern generators are already having a remarkable impact in robotics. Furthermore, given the latest advances in low-cost manufacture (a combination of 3D printing technology with of-the-shelf sensors and actuators), it will be very interesting to see what novel robot designs appear in the near future.

\vfill
\pagebreak

\phantomsection
\section*{Acknowledgments}
\addcontentsline{toc}{section}{Acknowledgments}

I would like to thank Professor Murray Shanahan for his constant enthusiasm in neuroscience and insightful advice for this project.

\phantomsection
\bibliographystyle{unsrt}
\bibliography{cpgreferences}


\end{document}